\documentclass[conference]{IEEEtran}
\IEEEoverridecommandlockouts
\usepackage{cite}
\usepackage{amsmath,amssymb,amsfonts}
\usepackage{algorithmic}
\usepackage{graphicx}
\usepackage{textcomp}
\usepackage{xcolor}
\usepackage{url}
\def\BibTeX{{\rm B\kern-.05em{\sc i\kern-.025em b}\kern-.08em
    T\kern-.1667em\lower.7ex\hbox{E}\kern-.125emX}}
\begin{document}

\title{Beyond General Prompts: Automated Prompt Refinement using Contrastive Class Alignment Scores for Disambiguating Objects in VLMs
}

\author{\IEEEauthorblockN{Lucas Choi}
\IEEEauthorblockA{\textit{Archbishop Mitty} \\
{lucasleechoi@gmail.com}
}
\and
\IEEEauthorblockN{Ross Greer}
\IEEEauthorblockA{\textit{University of California, Merced} \\
{rossgreer@ucmerced.edu}
}
}

\maketitle


\begin{abstract}
Vision-language models (VLMs) offer flexible object detection through natural language prompts but suffer from performance variability depending on prompt phrasing. In this paper, we introduce a method for automated prompt refinement using a novel metric called the Contrastive Class Alignment Score (CCAS), which ranks prompts based on their semantic alignment with a target object class while penalizing similarity to confounding classes. Our method generates diverse prompt candidates via a large language model and filters them through CCAS, computed using prompt embeddings from a sentence transformer. We evaluate our approach on challenging object categories, demonstrating that our automatic selection of high-precision prompts improves object detection accuracy without the need for additional model training or labeled data. This scalable and model-agnostic pipeline offers a principled alternative to manual prompt engineering for VLM-based detection systems.
\end{abstract}


\section{Introduction}

\begin{figure}
    \centering
    \includegraphics[width=.40\textwidth]{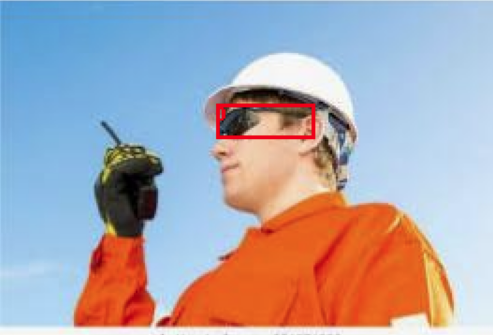}
    \caption{This is a sample detection from foundation VLM OWLv2 prompted with 'goggles'. The model mistakenly detected these sunglasses as goggles, which may have serious safety implications in a worksite monitoring task where safety goggles are important. As illustrated, certain prompts have ambiguity in their definitions, reflecting the many-object-encompassing aspect of natural language, but also resulting in poor precision detections and necessitating more descriptive prompts to ensure unintended objects are not mistakenly detected.}
    \label{fig:fault}
\end{figure}

Vision-language models (VLMs) have expanded object detection capabilities by replacing fixed class labels with open-ended natural language prompts. However, the performance of these models is highly sensitive to the phrasing and specificity of the prompts used\cite{zhou2022learning,dumpala2024sugarcrepe++, greer2025language, keskar2025evaluating}.
A generic prompt may not perform well compared to a carefully chosen descriptive prompt, yet on the other hand, a too descriptive prompt may cause a model to fail to detect the object \cite{du2024ipo}. In datasets where visual distinctions are critical—such as differentiating “safety goggles” from “glasses” or "sunglasses"—poor prompt choices can introduce ambiguity, such as in Figure \ref{fig:fault}, reducing both precision and recall \cite{geigle2024african, kim2024finer}. This sensitivity poses the challenge of systematically generating and selecting prompts that maximize detection accuracy while minimizing confusion with visually or semantically similar classes.

\begin{figure*}
    \centering
    \includegraphics[width=\textwidth]{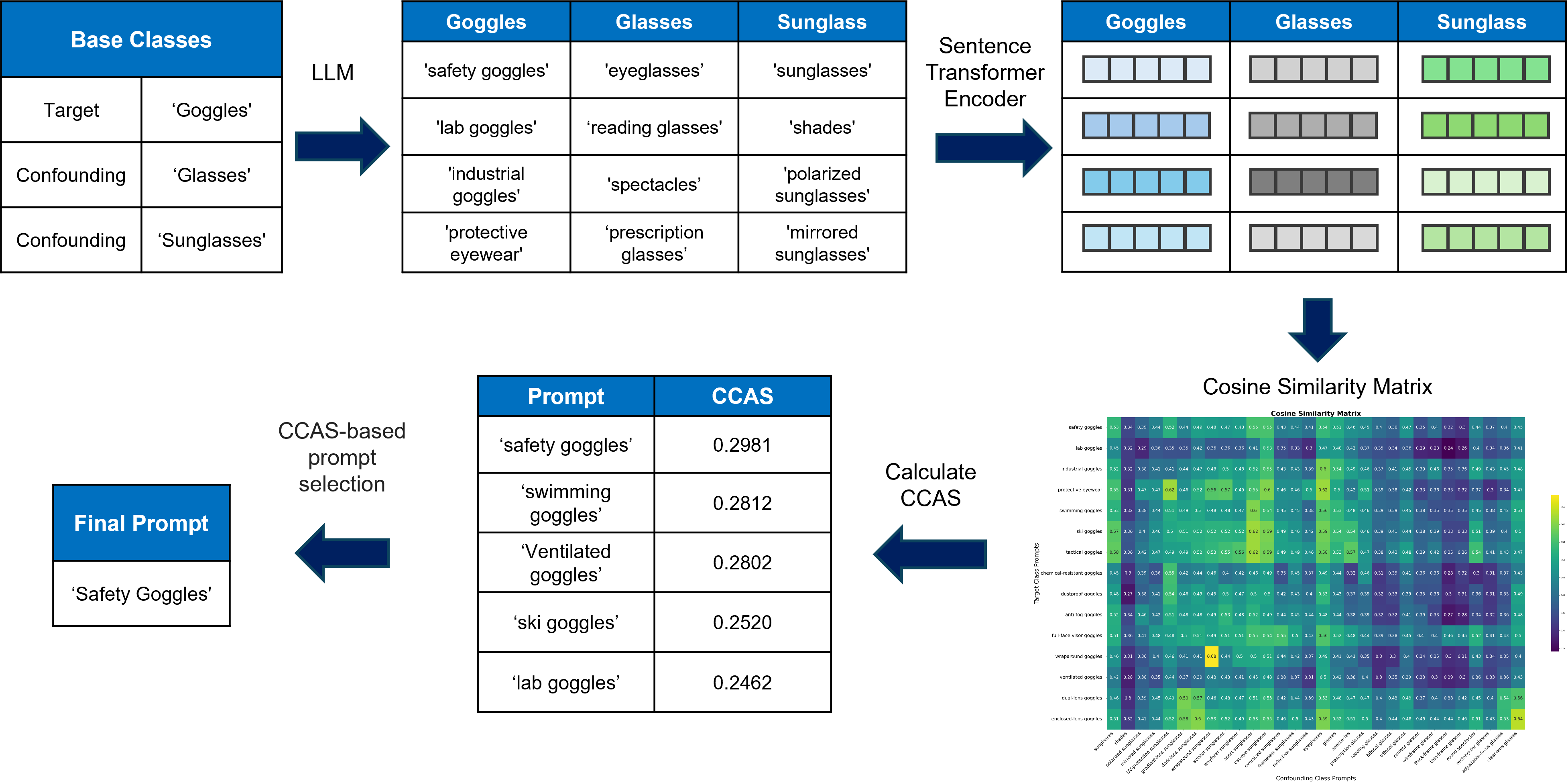}
    \caption{System Diagram of our proposed algorithmic pipeline for identifying an optimal high-precision prompt through our CCAS metric. The diagram utilizes goggles as the example classes and prompts, with only a sample of prompts being shown in the diagram. The complete pipeline and illustrated example are discussed in Algorithm and Experimental Evaluation.}
    \label{fig:Diagram}
\end{figure*}

In this research, we propose a method for algorithmically identifying high-precision natural language prompts for object detection in vision-language models using what we refer to as a contrastive class alignment score (CCAS). We propose a similarity-based prompt filtering pipeline that selects prompts most semantically aligned with a target object class while reducing confusion with similar but distinct classes. This process, as illustrated in Figure \ref{fig:Diagram}, provides a scalable alternative to manual prompt engineering and improves detection performance through prompt optimization.

\section{Related Research}
 
Existing methods for enhancing prompting for open-vocabulary object detection include fine-tuning an LLM that generates prompts, diversifying prompts to include both text and images \cite{medisetti2024litai, tian2024gaugetracker}, and visual input modification.

Avshalumov et al.
\cite{avshalumov2024reframing} defines a ``Reframing" method that uses feedback from detection models to finetune an LLM to optimize its queries, facilitating stronger detection performance. Du et al. \cite{du2022learning} introduce the DetPro method to learn prompt representations, which use background interpretation and separation of foreground elements during the training of the prompting model. By contrast, our method avoids LLM training or finetuning, using only an inference stage. 

The T-Rex2 model of Jiang et al. \cite{jiang2024t} extends the prompt reception of VLM-based detection models to accept input in the form of images and the combination of images and text; though the method achieves strong performance, we restrict our research problem to a true zero-shot (no prior exemplar images) setting, utilizing only abstract text prompts. 

Yang et al. \cite{yang2023fine} take an opposite approach to the enhancement problem, instead modifying the input image rather than the prompt via outline or mask elements, finding that blurring outside the target mask enhances localization. This method requires first the successful recognition of the target, then refinement of the object localization; our research is centered on the preliminary step of recognition. 

Recent advances also explore continuous prompt embeddings in approaches like CoOp by Zhou et al. \cite{zhou2022learning} and ProDA by Lu et al.\cite{lu2022prompt}, which fine-tune soft prompts for downstream vision tasks. These methods contrast with our discrete, interpretable phrase-level prompts but share the goal of adapting language to better match visual representations.

Additionally, interactive and human-in-the-loop methods have gained interest as viable strategies for dynamic prompt adaptation, especially in safety-critical applications. Studies such as SugarCrepe++ by Dumpala et al. \cite{dumpala2024sugarcrepe++} and FINER by Kim and Ji\cite{kim2024finer} focus on prompt robustness and sensitivity to semantic variation, underlining the importance of prompt phrasing in zero-shot VLM performance.

More complex approaches to prompt refinement exist for particular domains; for example, towards camouflaged-object recognition, Zhang et al. \cite{zhang2025camosam2} integrate motion and appearance cues to refine multiple prompts to the Segment Anything (SAM2) model \cite{kirillov2023segment}.Wu et al. present AttriPrompter \cite{wu2023zero, wu2024attriprompter}, an autoprompting pipeline using attribute generation, augmentation, and relevance sorting specific to the task of nuclei detection in histopathology images.

\section{Algorithm}

Input to our system includes a dataset, target class $T$, and a set of known confounding classes to the target $C: c_1, ..., c_m$. We generate $N$ possible versions and specifications of each class, both target and confounding, using a large language model (LLM). We refer to these LLM-augmented lists as $T: t_{1},...,t_{N}$ and $C_i: c_{i,1},...,c_{i,N}$, where $i$ refers to the index of the class in the original set $C$. We prompted the LLM: 
\begin{quote}
``Generate an extensive list of possible descriptions, synonyms, and detection-oriented prompts without negatives, limiting the prompt to a phrase, to detect the following base object classes with $N$ prompts per class, intended for use with a vision-language model: \textless class1\textgreater, \textless class2\textgreater, etc.''
\end{quote}
This prompt was chosen as short phrases are best for prompts as they disallow for too much specificity, which is non-optimal for detecting objects across the various environments in a typical image dataset. Additionally, negatives were discouraged as they only add confusion to the prompts, and it is unnecessary to prompt for what we do not want to detect. Finally, we specified the number of prompts generated per class, as the number is variable based on the base class name and the number of classes provided.


With the extensive list of prompts obtained for each class, including the base class name in the list, we then take the embeddings of each prompt using a sentence transformer to make comparisons of semantic meaning, specifically between the target and the confusion classes. This is to help narrow the obtained prompts and reduce the overlap in generated prompts. To achieve this, we compute the cosine similarity between every pair of embeddings between a target class and its confounding classes, constructing a similarity matrix. 

With the target class on the y-axis, we take the average similarity from each row and compute the CCA scores for the precision of each prompt of the target class. This is done through the two following equations, investigating which of the two results in higher accuracy prompts.

\begin{equation}
CCAS_{avg}(t_{i}) = \cos(\vec{t_{i}}, \vec{T}) - \frac{1}{NM}
\sum_{m=1}^{M} \sum_{k=1}^{N} \cos(\vec{t_{i}}, \vec{c}_{m,k})
\end{equation}

\begin{equation}
CCAS_{max}(t_{i}) = \cos(\vec{t_{i}}, \vec{T}) - \max_{m,k}\cos(\vec{t_{i}}, \vec{c}_{m,k})
\end{equation}

 where  
 \begin{itemize}
    \item $\vec{t_{i,j}}$: Embedding vector of the $j$-th candidate prompt for target class $i$
    \item $\vec{t_i}$: Embedding vector of the $i$-th target base class name
    \item $\vec{c}_{m,k}$: Embedding vector of the $k$-th prompt of the $m$-th confounding class
    \item $\cos(\vec{a}, \vec{b})$: Cosine similarity between vectors $\vec{a}$ and $\vec{b}$
    \item $M$: Total number of confounding classes
    \item $N$: Total number of prompts per class
\end{itemize}

By ranking the prompts with their score, we can take the top target prompts to remove vague and general prompts or prompts that overlap with confusion classes.

\section{Experimental Evaluation}

To evaluate the proposed algorithm, we used two datasets: the Safety Goggles Computer Vision Project dataset\footnote{https://universe.roboflow.com/database-sjrvw/safety-goggles} with the target class of `goggles' and confusion classes of `glasses' and `sunglasses', and the Self-Driving Cars Computer Vision Project dataset \cite{self-driving-cars-lfjou_dataset}, with `stop' as the target class and `red light' and `speed limit' as the confounding classes. Confounding classes were selected due to their frequent appearance in the dataset and similarity to the target class. We set $N$ as 15 and 25 for the goggles and stop sign tasks, respectively.

\begin{figure*}
    \centering
    \includegraphics[width=.98\textwidth]{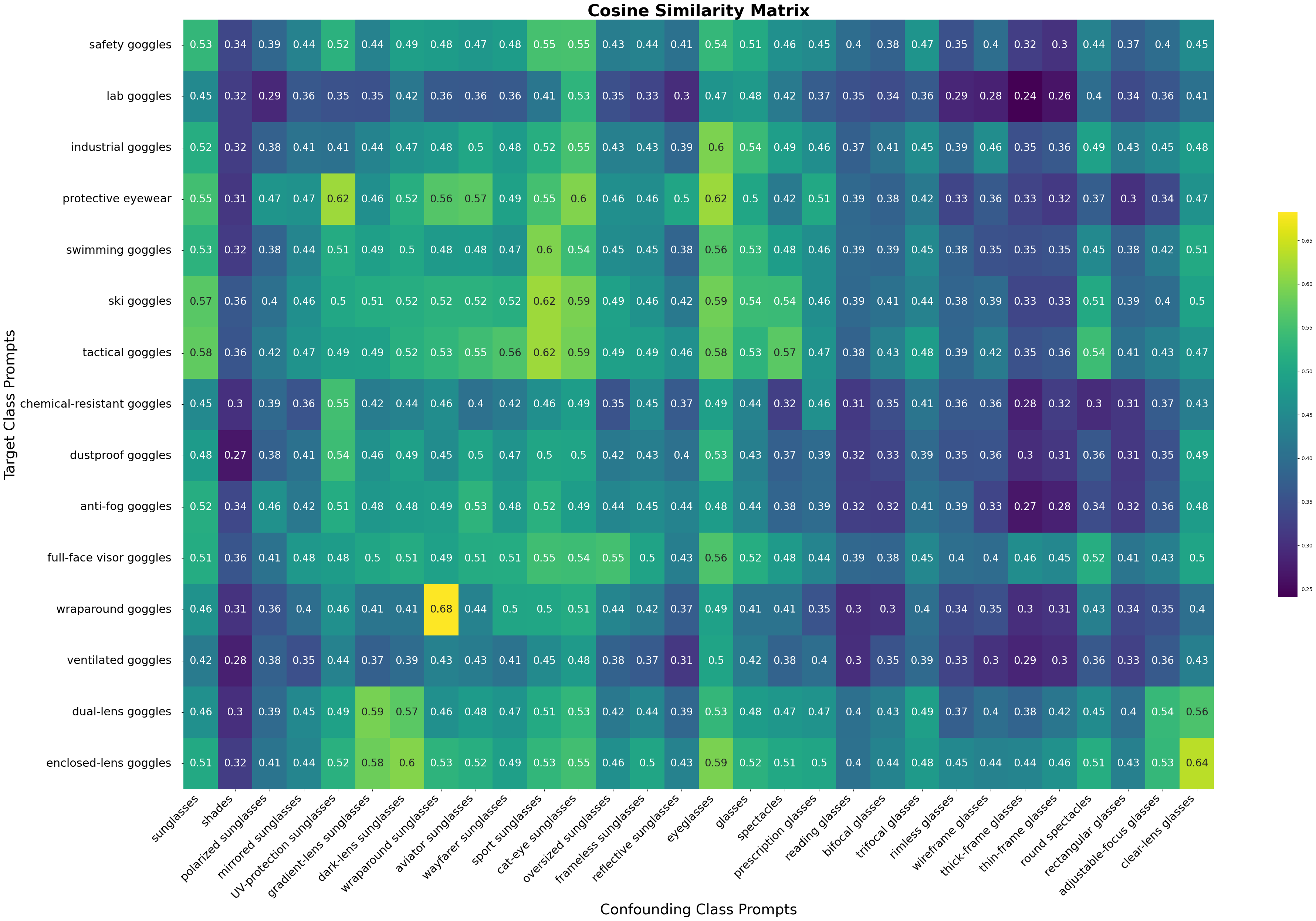}
    \caption{Similarity Matrix of the Goggle Detection task prompts, specifically between the `goggle', `glasses', and `sunglasses' classes. The y-axis consists of the target class prompts, while the x-axis consists of all of the prompts from the confounding classes.}
    \label{fig:heatmap}
\end{figure*}

GPT-4o\cite{hurst2024gpt} was chosen as the LLM for prompt generation. We then utilized the sentence transformer, all-MiniLM-L6-v2, the fine-tuned version of Minilm\cite{wang2020minilm}, to create embedding. With these embeddings, we generated the similarity matrix of cosine similarities as shown in Figure \ref{fig:heatmap} and computed the CCA scores through both equations of each target class prompt as shown in Table \ref{tab:CCAS_avg} and Table \ref{tab:CCAS_max}. The examples shown in the figures and tables are both of the goggle detection task.

\begin{table}
\centering
\caption{Prompt variations and their CCA scores averages, sorted in descending order.}
\begin{tabular}{l|c}
Prompt & $CCAS_{avg}$ \\
\hline
\hline
swimming goggles & 0.4294 \\
safety goggles & 0.4128 \\
lab goggles & 0.4098 \\
ventilated goggles & 0.4019 \\
ski goggles & 0.4004 \\
tactical goggles & 0.3794 \\
industrial goggles & 0.3614 \\
dustproof goggles & 0.3568 \\
wraparound goggles & 0.3563 \\
anti-fog goggles & 0.3194 \\
dual-lens goggles & 0.3135 \\
chemical-resistant goggles & 0.2966 \\
full-face visor goggles & 0.2917 \\
enclosed-lens goggles & 0.2878 \\
protective eyewear & 0.0621 \\
\end{tabular}
\label{tab:CCAS_avg}
\end{table}

\begin{table}
\centering
\caption{Prompt variations and their CCA scores maxes, sorted in descending order.}
\begin{tabular}{l|c}
Prompt & $CCAS_{max}$ \\
\hline
\hline
safety goggles & 0.2981 \\
swimming goggles & 0.2812 \\
ventilated goggles & 0.2802 \\
ski goggles & 0.2520 \\
lab goggles & 0.2462 \\
tactical goggles & 0.2413 \\
dustproof goggles & 0.2229 \\
industrial goggles & 0.2140 \\
anti-fog goggles & 0.2122 \\
full-face visor goggles & 0.2016 \\
dual-lens goggles & 0.1794 \\
enclosed-lens goggles & 0.1429 \\
chemical-resistant goggles & 0.1403 \\
wraparound goggles & 0.0788 \\
protective eyewear & -0.1016 \\
\end{tabular}

\label{tab:CCAS_max}
\end{table}


We used OWLv2\cite{minderer2023scaling} as the vision language model for zero-shot object detection to benchmark this method without having variability in the training. We evaluate the differences in performance based on how many of the top-scoring prompts from each CCAS method we give to the model as well as if we prompted only the base class name as shown in Table \ref{tab:GoggleAP} and \ref{tab:SignsAP}, with average precision (AP) as the metric of evaluation. We use the class name provided directly by the dataset as baseline, in evaluating a hypothetical fully-automated pipeline without human intervention. Such intervention would bias the experiment not only in the selection of base class, but also in the filtering of obviously incorrect prompts (e.g. swimming goggles and ski goggles for the lab safety goggles setting); we allow these errors to propagate to show the robustness of the method when considering the top-$n$ prompts during our experiments.  

\begin{table}
\centering
\caption{Comparison of prompt configurations and their corresponding average precisions on Safety Goggles Detection}
\begin{tabular}{c|c|c}
Prompt Configuration & $CCAS_{avg}$ AP & $CCAS_{max}$ AP \\
\hline \hline
Baseline (``goggles'') & 0.2555 & 0.2555\\ 
\hline
$CCAS_{Top 1}$ & 0.3559 & 0.5415\\
\hline 
$CCAS_{Top 3}$ & 0.5108 & 0.5279\\
\hline 
$CCAS_{Top 5}$ & 0.5049 & 0.5049\\
\hline 
$CCAS_{Top N}$ & 0.4343 & 0.4343\\
\hline 

\end{tabular}
\label{tab:GoggleAP}
\end{table}

\begin{table}
\centering
\caption{Comparison of prompt configurations and their corresponding average precisions on Stop Sign Detection}
\begin{tabular}{c|c|c}
Prompt Configuration & $CCAS_{avg}$ AP & $CCAS_{max}$ AP \\
\hline \hline
Baseline (``stop") & 0.006858 & 0.006858\\ 
\hline
$CCAS_{Top 1}$ & 0.3835 & 0.3045\\
\hline 
$CCAS_{Top 3}$ & 0.3118 & 0.2991\\
\hline 
$CCAS_{Top 5}$ & 0.1808 & 0.1856\\
\hline 
$CCAS_{Top N}$ & 0.1939 & 0.1939\\
\hline 
\end{tabular}
\label{tab:SignsAP}
\end{table}

\section{Discussion}
The OWLv2 evaluation results demonstrate that fewer high-scoring prompts result in higher average precision. Specifically for the goggle evaluation, the $CCAS_{max}$ had consistently better performance for the top 3 and top 1 prompts. However, for the stop sign task, the $CCAS_{avg}$ performed better, demonstrating that both are plausible methods to form top-contender prompts.

As shown with the top 5 performance, both methods of averages and maxes output a similar list. However, as we reduce the number of top prompts taken, we can notice the effects of small changes in the orders of the prompts, as `ventilated goggles' scores higher than `lab goggles', and `safety goggles' scores higher than `swimming goggles. 

Additionally, it is best to choose the top single prompt from the $CCAS$ method as average precision notably decreases with the addition of more prompts. Although this may increase recall, it contributes to the same problem of ambiguity, where there is a higher chance of detecting unintended objects with a wider variety of prompts.

Based on the amount of ambiguity between the confounding classes and the target class, there exist instances when the base object class name is distinguishable enough to perform the best. In this case, we would choose the baseline prompt over the CCAS top-scoring prompts. On the other hand, with enough ambiguity, the CCAS prompts help to discern specifically for the desired object.

Literature review suggests that measuring ambiguity in language has been investigated as an interesting niche problem of linguistics \cite{dumitrache2015crowdtruth, kiyavitskaya2008requirements, ceccato2004ambiguity}, followed by early application in object description for computer vision \cite{mao2016generation}, and we expect this research area to be increasingly relevant as more prompt-driven models become effective at zero-shot performance on tasks relevant to an expanding set of domain applications \cite{greer2024towards, bossen2025can}. 

While our results demonstrate improvements, the method’s effectiveness is inherently influenced by the diversity and quality of the LLM-generated prompt pool. Future extensions could incorporate user feedback or reinforcement learning strategies to iteratively refine prompt sets based on real-world model behavior.

The use of discrete, human-readable prompts scored through contrastive alignment creates an interpretable layer between user input and model output. Our pipeline allows users to understand why certain prompts performed better based on their semantic distance from confounding classes. This transparency is particularly valuable in domains like healthcare or industrial safety, where black-box systems face resistance. As such, CCAS does more than improve performance—it enhances the trustworthiness of zero-shot detection pipelines.


\section{Concluding Remarks}

In this research, we present a method for the automatic refinement of prompts in the presence of confounding classes. The method does not require model finetuning, leveraging cosine distances between positive and negative paired words to identify prompts that best describe a target class with minimal overlap to additional classes imagined by an LLM in the system pipeline. This method may be applied in zero-shot for the detection of objects where distinct recognition between objects with similar attributes is important, such as safety-based domains, where an object improperly identified or improperly worn by a user may have serious consequences \cite{choi2024evaluating, choi2024evaluatingmotorcycle, greer2024driver, shriram2025towards}.
As the utilization of vision-language models increases across many application domains, automated prompt refinement presents an opportunity for improved model accuracy to enable downstream perception, planning, and control tasks in intelligent and autonomous systems.  
 
{\small
\bibliographystyle{ieeetr}
\bibliography{refs}
}

\end{document}